\documentclass{article}

\usepackage[square,numbers]{natbib}
\usepackage[final, nonatbib]{neurips_2020}

\usepackage[utf8]{inputenc} % allow utf-8 input
\usepackage[T1]{fontenc}    % use 8-bit T1 fonts
\usepackage{hyperref}       % hyperlinks
\usepackage{url}            % simple URL typesetting
\usepackage{booktabs}       % professional-quality tables
\usepackage{amsfonts}       % blackboard math symbols
\usepackage{nicefrac}       % compact symbols for 1/2, etc.
\usepackage{microtype}      % microtypography
\usepackage{graphicx}
\usepackage{subcaption}
\usepackage{sidecap}

\usepackage{wrapfig}
\usepackage{booktabs}       % professional-quality tables
\usepackage{multirow}
\usepackage{color}
\usepackage{colortbl}
\definecolor{Gray}{gray}{0.9}
\newcommand{\CC}{\cellcolor{Gray}}

\title{Enhancing Poaching Predictions for Under-Resourced Wildlife Conservation Parks Using Remote Sensing Imagery}

\author{
  Rachel Guo$^1$\thanks{Contact authors \texttt{rguo@college.harvard.edu} and \texttt{lily\_xu@g.harvard.edu}.}, Lily Xu$^1$, Drew Cronin$^2$, Francis Okeke$^3$, Andrew Plumptre$^4$, Milind Tambe$^1$  
  \\
  $^1$Harvard University, $^2$North Carolina Zoo, $^3$WCS Nigeria, $^4$Key Biodiversity Secretariat \\
}

\begin{document}

\maketitle

\begin{abstract}
 Illegal wildlife poaching is driving the loss of biodiversity. To combat poaching, rangers patrol expansive protected areas for illegal poaching activity. However, rangers often cannot comprehensively search such large parks. Thus, the Protection Assistant for Wildlife Security (PAWS) was introduced as a machine learning approach to help identify the areas with highest poaching risk. As PAWS is deployed to parks around the world, we recognized that many parks have limited resources for data collection and therefore have scarce feature sets. To ensure under-resourced parks have access to meaningful poaching predictions, we introduce the use of publicly available remote sensing data to extract features for parks. By employing this data from Google Earth Engine, we also incorporate previously unavailable dynamic data to enrich predictions with seasonal trends. We automate the entire data-to-deployment pipeline and find that, with only using publicly available data, we recuperate prediction performance comparable to predictions made using features manually computed by park specialists. We conclude that the inclusion of satellite imagery creates a robust system through which parks of any resource level can benefit from poaching risks for years to come.
\end{abstract}

\section{Introduction}
\label{sec:intro}

\begin{wrapfigure}{r}{.45\textwidth}
  \centering
    \includegraphics[width=\linewidth]{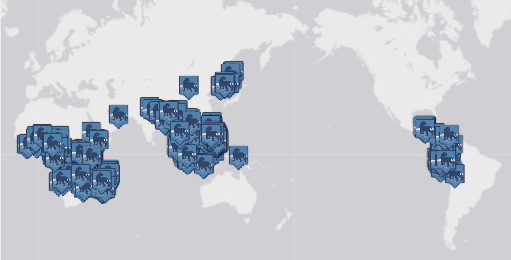}
  \caption{SMART is used by 800 protected areas around the world, many of which have limited access to geospatial data.}
  \label{fig:smart_map}
\end{wrapfigure}

Illegal poaching of wildlife threatens endangered species, driving the loss of biodiversity and contributing to the climate crisis \cite{pires2016illegal}. To deter poaching, rangers search protected areas and remove snares laid out to trap animals \cite{o2018experimental}. However, conservation parks have a limited number of rangers to search parks that scale to thousands of square kilometers. Thus, the Protection Assistant for Wildlife Security (PAWS) has been developed as a machine learning approach to predict areas of highest poaching risk based on historical poaching patterns and geospatial features. PAWS is currently being integrated with SMART (Fig.~\ref{fig:smart_map}), a leading conservation software, to become deployed to 800 parks around the world \cite{xu}.

In a series of alpha tests of the PAWS integration with SMART involving over 20 parks, some park managers reported nonsensical predictions because they had just a single feature: park boundary. We recognized that an AI system would be useful only if we helped address these challenges of unavailable features, but seeing that some features require expensive aerial surveys to compute and GIS specialists to create \cite{rowcliffe2008estimating}, we were limited to processes that required virtually no effort or technical expertise from the parks. 

In order to make the PAWS system resilient, scalable, and deployable to parks around the world of varying resources, we introduce an automatic pipeline into PAWS that extracts remote sensing data from Google Earth Engine (GEE), which ensures every park has access to the same abundant feature set. We discover that we recover almost all predictive performance using only GEE features on three test parks---Murchison Falls National Park (MFNP) and Queen Elizabeth National Park (QENP) in Uganda and Cross River National Park (CRNP) in Nigeria. We also find that dynamic data from GEE prove useful in predictive performance and that we can recreate some features provided by parks exactly using GEE imagery. This makes for a robust system on which PAWS can make poaching predictions virtually anywhere in the world for decades to come.

\paragraph{Related Work}

% PAWS was initially incepted as a predictive modelling approach to anti-poaching, and has since been iterated upon as it is deployed around the world \cite{gholami}. While remote sensing data has been used in several applications in past decades \cite{gee-applications}, in the space of machine learning for wildlife protection, the use of remote sensing has been limited. 

PAWS was initially incepted as a predictive modelling approach to anti-poaching \cite{yang}, and has since been iterated upon as it continues to be deployed around the world \cite{xu,fang2016deploying,gholami2017taking}. 

While remote sensing data has been used in several applications in past decades \cite{gee-applications}, from applications in coral reefs \cite{garza-perez}, to landslide activity \cite{casson}, to crop mapping \cite{shelestov}, in the space of machine learning for wildlife protection, the use of remote sensing has been limited. 

We use Google Earth Engine to source remote sensing data because GEE is publicly available, easy to use via an API, and includes a vast global selection of geospatial and climactic data \cite{gee-paper}. These GEE remote sensing datasets come from a vast catalog of remote sensing geospatial datasets available on a global scale in real time surveyed by institutions such as NASA, USGS, and NOAA. 

\section{Methods}
\paragraph{Extracting remote sensing data}

We create an automated pipeline that extracts remote sensing data from Google Earth Engine (GEE). We identified 13 features from GEE as potentially useful for poaching predictions \cite{critchlow2015spatiotemporal}. Static features we include are land cover, rivers, surface water, flow accumulation, elevation, slope, aspect, and drainage direction, and dynamic features we include are temperature, precipitation, net primary productivity, gross primary productivity, aerosol, and cirrus atmospheric content.

\paragraph{Preprocessing}

\begin{figure}[!ht]
  \centering
  \begin{subfigure}[t]{0.236\linewidth}
    \centering
    \includegraphics[height=2.6cm]{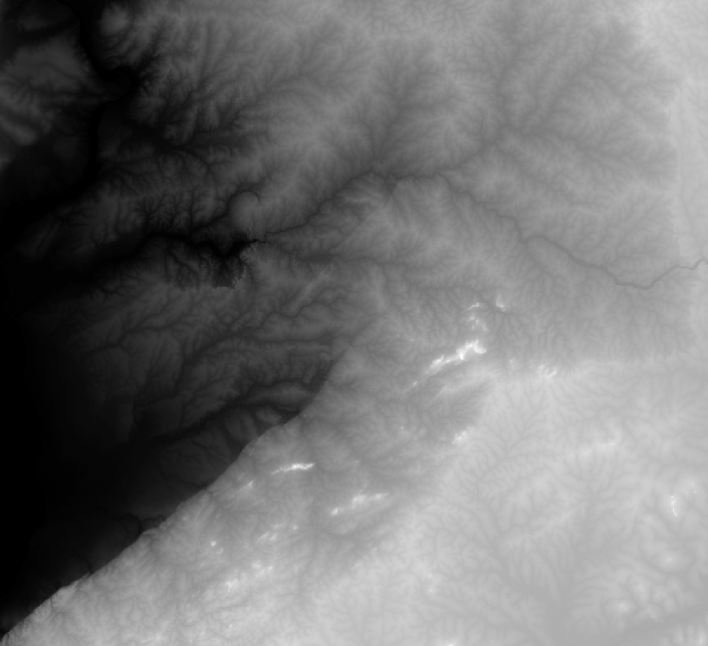}
    \caption{Raw DEM from GEE}
  \end{subfigure}~
  \begin{subfigure}[t]{0.236\linewidth}
  \centering
    \includegraphics[height=2.6cm]{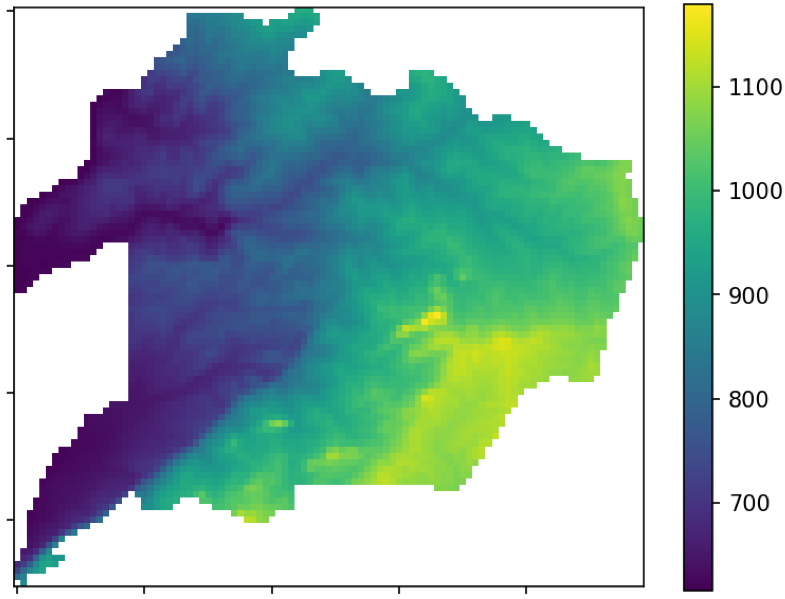}
    \caption{Rasterized elevation}
  \end{subfigure}~
  \begin{subfigure}[t]{0.236\linewidth}
  \centering
    \includegraphics[height=2.6cm]{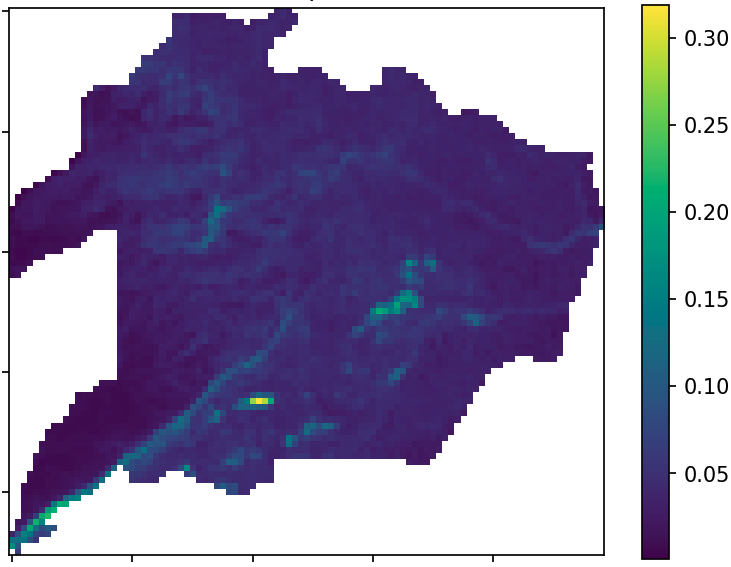}
    \caption{Rasterized slope}
  \end{subfigure}~
  \begin{subfigure}[t]{0.236\linewidth}
  \centering
    \includegraphics[height=2.6cm]{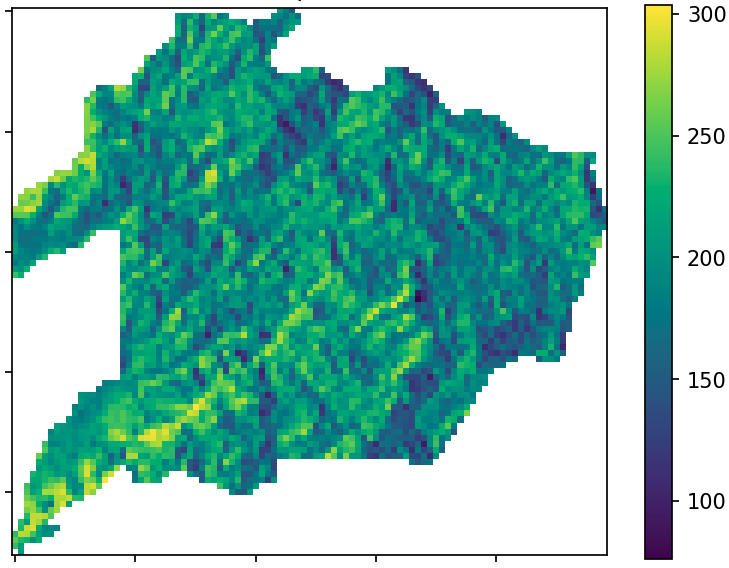}
    \caption{Rasterized aspect}
  \end{subfigure}
  \caption{Raw digital elevation model (DEM) TIF extracted from GEE (90 m) discretized into $1 \times 1$ km cells in MFNP boundary (1000 m). We also compute slope and aspect from the DEM.}
  \label{fig:elevation_tif_to_png}
\end{figure}

We discretize the park boundary into $1 \times 1$ km cells for a 1000 m spatial resolution. We rasterize each feature into TIF files that match the spatial resolution of the park. For shapefiles provided by parks, which include roads, rivers, villages, animal density, elevation, patrol posts, and the park boundary, each $1 \times 1$ km cell represents distance to the nearest feature. High-resolution imagery from GEE is rasterized into a lower resolution of 1000 m. For dynamic data, we discretize time into three-month intervals to account for seasonal poaching patterns historically found.

\begin{figure}[!ht]
  \centering
  \begin{subfigure}[t]{0.23\linewidth}
    \centering
    \includegraphics[height=3cm]{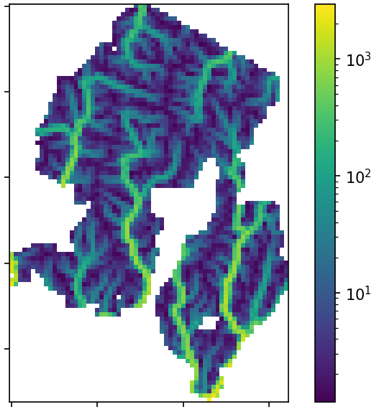}
    \caption{flow accumulation}
  \end{subfigure} 
  \begin{subfigure}[t]{0.23\linewidth}
  \centering
    \includegraphics[height=3cm]{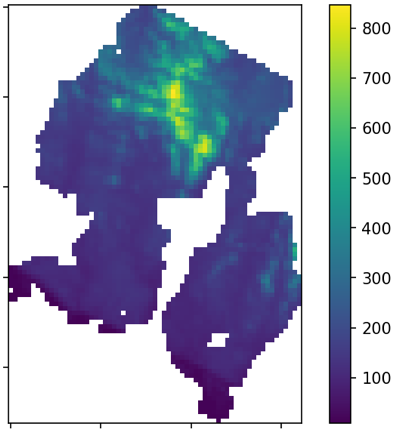}
    \caption{elevation}
  \end{subfigure}
  \begin{subfigure}[t]{0.23\linewidth}
  \centering
    \includegraphics[height=3cm]{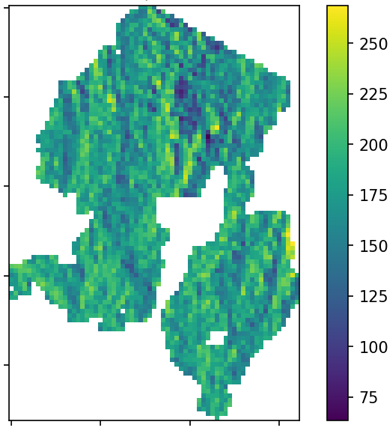}
    \caption{aspect}
  \end{subfigure}
  \begin{subfigure}[t]{0.23\linewidth}
  \centering
    \includegraphics[height=3cm]{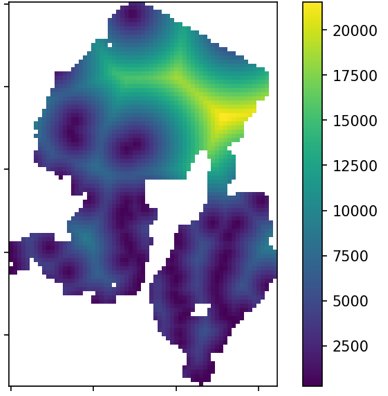}
    \caption{surface water distance}
  \end{subfigure} 
  \begin{subfigure}[t]{0.23\linewidth}
  \centering
    \includegraphics[height=3cm]{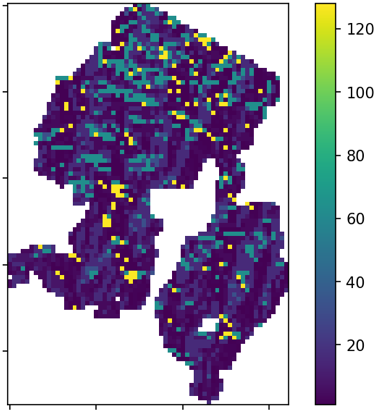}
    \caption{drainage direction}
  \end{subfigure} 
  \begin{subfigure}[t]{0.23\linewidth}
  \centering
    \includegraphics[height=3cm]{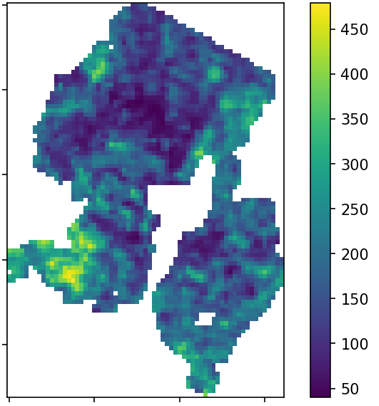}
    \caption{GPP (June 2019)}
  \end{subfigure} 
  \begin{subfigure}[t]{0.23\linewidth}
  \centering
    \includegraphics[height=3cm]{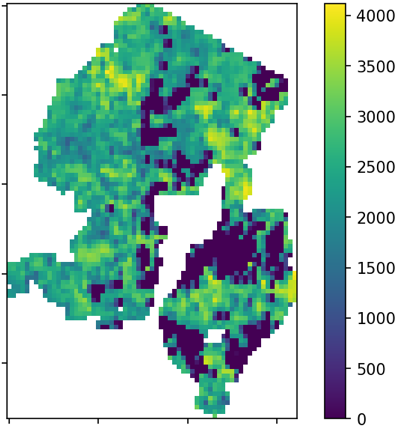}
    \caption{NPP (June 2019)}
  \end{subfigure} 
  \begin{subfigure}[t]{0.24\linewidth}
  \centering
    \includegraphics[height=3cm]{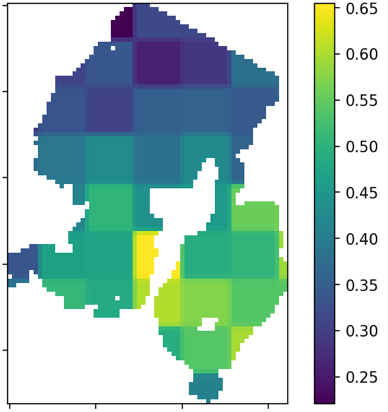}
    \caption{precip. (Feb~2018)}
  \end{subfigure}
  
  \begin{subfigure}[t]{0.23\linewidth}
  \centering
    \includegraphics[height=3cm]{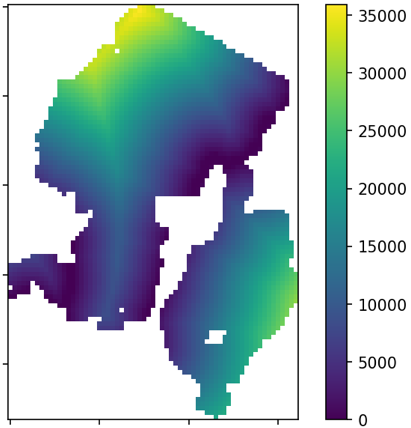}
    \caption{roads}
  \end{subfigure}
  \begin{subfigure}[t]{0.23\linewidth}
  \centering
    \includegraphics[height=3cm]{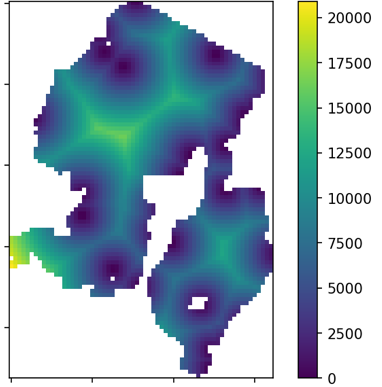}
    \caption{villages}
  \end{subfigure}
  \begin{subfigure}[t]{0.23\linewidth}
  \centering
    \includegraphics[height=3cm]{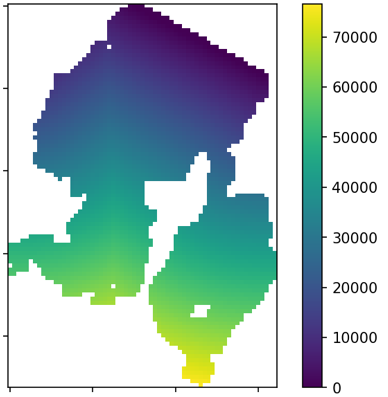}
    \caption{forest reserve}
  \end{subfigure}
  \begin{subfigure}[t]{0.23\linewidth}
  \centering
    \includegraphics[height=3cm]{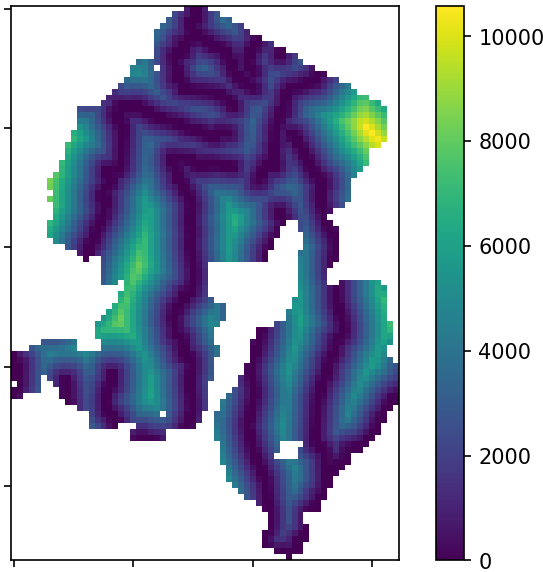}
    \caption{rivers}
  \end{subfigure}
  \caption{Preprocessed features extracted from GEE (first two rows) vs. features provided by CRNP park (bottom row). Drainage direction is categorical, while net primary production (NPP), gross primary production (GPP), and precipitation are dynamic features averaged over each month. Patterns in GEE's flow accumulation (d) are similar to the park's river feature (l).}
  \label{fig:gee}
 
\end{figure}

\paragraph{Making predictions}
Our labels are a binary indicator of past illegal activity for each cell. Past illegal activity data is incomplete as rangers are unable to search the entire park; one of our key challenges is to predict risk in areas not previously patrolled.

Another challenge is that negative labels are unreliable, as rangers may not detect snares during a patrol even though one is present. To help address this uncertainty, we use iWare-E (imperfect-observation aWare Ensemble model), a bagging ensemble of decision trees as weak learners, training each weak learner from different bins of input data based on current patrol effort \cite{gholami}. We make predictions on different thresholds of current patrol effort to help parks plan patrols based on the amount of effort they can expend. 

% \begin{figure}[h!]
%   \centering
%   \begin{subfigure}[t]{0.4\linewidth}
%     \centering
%     \includegraphics[height=3cm,width=3cm]{img/patrol_effort.png}
%     \caption{Past patrol effort}
%     \label{fig:past_patrol}
%   \end{subfigure}
%   \begin{subfigure}[t]{0.4\linewidth}
%   \centering
%     \includegraphics[height=3cm,width=3cm]{img/illegal_activity.png}
%     \caption{Past illegal activity}
%     \label{fig:past_illegal}
%   \end{subfigure}
%   \caption{QENP historical patrol effort and illegal activity (2013--2015)}
%   \label{fig:past_vis}
% \end{figure}

\section{Results}
We evaluate prediction accuracies for cases in which the feature set consists of: (i)~only features provided by parks, (ii)~only GEE features, and (iii)~all features (both park and GEE features). We present the performance, assessed by AUC, in Table~\ref{AUC-table} for parks MFNP, QENP, and CRNP. For each test year, we train our model on data from three years prior and test on the fourth year, which is listed.

% % risk maps
% \begin{figure}[h!]
%   \centering
%   \begin{subfigure}[t]{0.3\linewidth}
%     \centering
%     \includegraphics[height=3cm]{img/MFNP_risk_dt_combined.png}
%     \caption{MFNP (2017)}
%   \end{subfigure} \quad
%   \begin{subfigure}[t]{0.3\linewidth}
%   \centering
%     \includegraphics[height=3cm]{img/QENP_risk_dt_combined.png}
%     \caption{QENP (2016)}
%   \end{subfigure} \quad
%   \begin{subfigure}[t]{0.3\linewidth}
%   \centering
%     \includegraphics[height=3cm]{img/CRNP_risk.png}
%     \caption{CRNP (2019)}
%   \end{subfigure}
%   \caption{Predicted poaching risks for guide rangers to search areas with highest poaching risk (dark red).}
%   \label{fig:risk_maps}
% \end{figure}

\begin{figure}[!ht]
  \centering
  \begin{subfigure}[t]{0.23\linewidth}
    \centering
    \includegraphics[height=3cm]{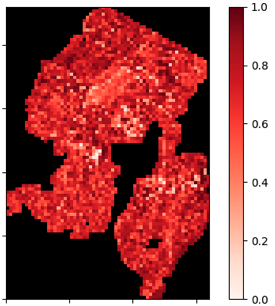}
    \caption{Baseline (2 feat.)}
  \end{subfigure}
  \centering
  \begin{subfigure}[t]{0.24\linewidth}
    \centering
    \includegraphics[height=3cm]{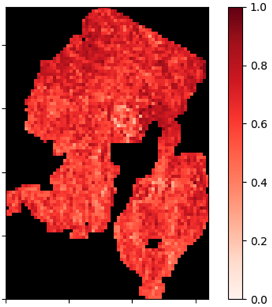}
    \caption{2 feat. + GEE (15 feat.)}
  \end{subfigure}
  \centering
  \begin{subfigure}[t]{0.23\linewidth}
    \centering
    \includegraphics[height=3cm]{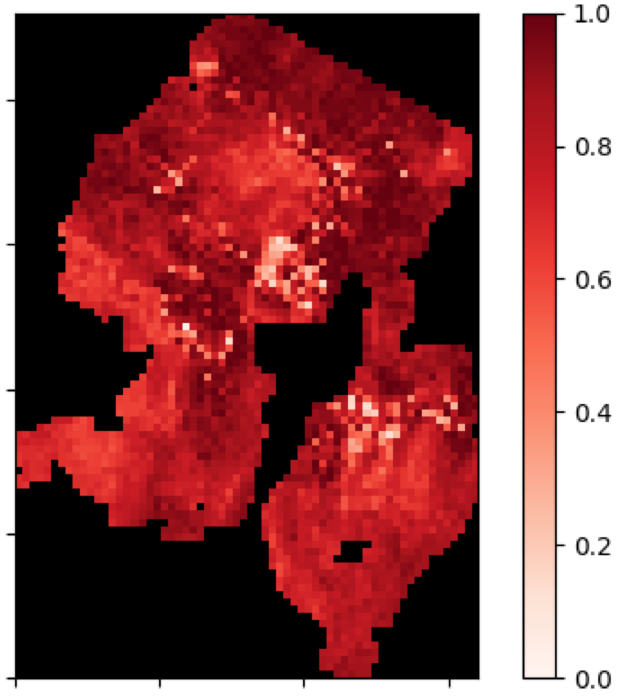}
    \caption{Baseline (11 feat.)}
  \end{subfigure}
%   \begin{subfigure}[t]{0.3\linewidth}
%   \centering
%     \includegraphics[height=3cm]{img/risk_comparisons/gee.png}
%     \caption{Only GEE}
%   \end{subfigure} \quad
  \begin{subfigure}[t]{0.23\linewidth}
  \centering
    \includegraphics[height=3cm]{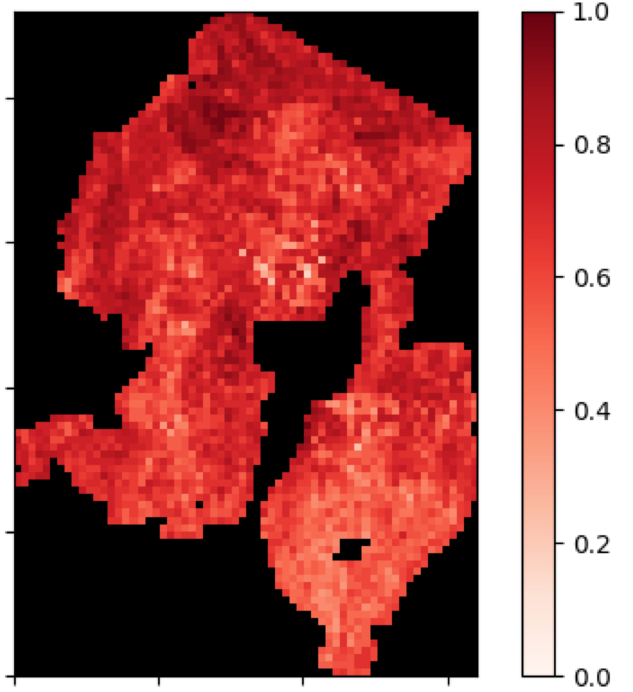}
    \caption{All Features (24 feat.)}
  \end{subfigure}
  \caption{Comparison of predicted poaching risks for CRNP with only park's features (Baseline), only GEE features (Only GEE), and park's and GEE features combined (All Features). We consider (a) only two features from the park (rivers and roads), (b) park's two features + 13 GEE features, (c) 11 park features without GEE, (d) 11 park features + 13 GEE features. There is considerable noise in (a) and (c), and adding GEE features leads to smoother predictions in (b) and (d). Rangers can then reliably use our poaching predictions to efficiently search parks.}
  \label{fig:risk_comparisons}
\end{figure}

\emph{We achieve nearly the same AUCs using satellite imagery, without any of the several customized features from parks.} As shown in Fig.~\ref{fig:risk_comparisons}, using only park's features versus both park's features and GEE lead to predictions in different areas of the park, and adding GEE reduces noise and yields smoother predictions. Although having only two features (Fig.~\ref{fig:risk_comparisons}a) may seem like an exaggeration, as mentioned in Sec.~\ref{sec:intro}, some parks using PAWS in practice attempted to make predictions with only park boundary as a feature. Thus, the scenario in Fig.~\ref{fig:risk_comparisons}a is realistic, particularly for under-resourced parks, and we show in Fig.~\ref{fig:risk_comparisons}b that including GEE features results in less noisy and more sensible predictions. Additionally, even feature-rich parks that have many curated features (considered in Fig.~\ref{fig:risk_comparisons}c which uses 11 features) would benefit from the improved predictions offered by the inclusion of GEE features (Fig.~\ref{fig:risk_comparisons}d). 

Traces of patterns in the features in Fig.~\ref{fig:gee} are evident in the respective risk maps. For example, NPP patterns are evident in the southeast portion of the park, suggesting that poaching patterns may be correlated with NPP. We noticed some dynamic features that were highly-ranked in feature importance had low spatial resolution. In fact, many of these dynamic features, such as aerosol content in the atmosphere, would seem to have no effect on poaching patterns. However, many of these dynamic features are correlated, and since dynamic data have cyclic, seasonal patterns, this tells us that dynamic data serves as a good proxy for learning patterns in other features when taken in tandem. We further note that we are essentially able to recreate some features that parks provided, such as river shapefiles, using GEE remote sensing data. 

\begin{SCtable}
    \caption{AUCs for predictions based on MFNP's 21 features, QENP's 19 features, and CRNP's 11 features (Baseline); only GEE features; and all features combined for each test year. While we have access to more historical data for MFNP and QENP, we have more recent data available for CRNP.}
    \label{AUC-table}
    \centering
    % \begin{tabular}{c|g|gggc@{}}
    \begin{tabular}{c|c|ccc}
    
        \toprule
        \multicolumn{2}{c}{}& Baseline & GEE & All Features \\
        \midrule
        \multirow{5}{*}{\rotatebox[origin=c]{90}{MFNP}}
        &$2014$ & $0.707$ & $0.696$ & $\textbf{0.714}$ \\ 
        &$2015$ & $\textbf{0.678}$ & $0.652$ & $0.671$ \\ 
        &$2016$ & $0.678$ & $0.647$ & $\textbf{0.681}$ \\ 
        &$2017$ & $0.683$ & $0.665$ & $\textbf{0.688}$ \\ 
        & \CC Avg &
        \CC $0.687$ & \CC $0.665$ & \CC \textbf{0.689} \\
        \midrule
        
        \multirow{4}{*}{\rotatebox[origin=c]{90}{QENP}}
        &$2014$ & $0.710$ & $0.697$ & $\textbf{0.716}$ \\
        &$2015$ & $0.710$ & $\textbf{0.722}$ & $0.720$ \\ 
        &$2016$ & $0.715$ & $0.699$ & $\textbf{0.716}$ \\
        & \CC Avg & \CC $0.712$ & \CC $0.706$ & \CC $\textbf{0.717}$ \\ 
        \midrule
    
        \multirow{3}{*}{\rotatebox[origin=c]{90}{CRNP}}
        & $2018$ & $0.655$ & $\textbf{0.693}$ & $0.670$ \\ 
        & $2019$ & $\textbf{0.741}$ & $0.727$ & $\textbf{0.741}$ \\
        & \CC Avg & \CC $0.698$ & \CC $\textbf{0.710}$ & \CC $0.706$ \\ 

        \bottomrule
    \end{tabular}
\end{SCtable}

\section{Conclusion and Future Work}

Under-resourced parks without the means --- expertise, equipment, and time --- to create their own features can benefit greatly from extracting features from publicly available satellite imagery to generate poaching predictions. As PAWS continues to be improved and deployed to parks around the world, we also consider next steps such as more explainable AI as we strive to better understand exactly what features are associated with higher poaching risks in different areas of parks. We aim to employ CNNs with a decision-focused learning approach to further improve upon our predictions. We are interested in expanding our feature set further by exploring additional publicly available data products such as OpenStreetMap, as well as creating our own data products from scratch using vision.

\section*{Acknowledgments}
% \paragraph{Acknowledgments}

CRNP data courtesy of WCS Nigeria and MFNP/QENP data courtesy of the Uganda Wildlife Authority. Thank you to the rangers who are doing critical work to protect wildlife on the ground, the NC Zoo for the collaboration, and SMART Partnership for their conservation leadership.

% \section*{References}

\bibliography{bibfile}

\end{document}